\documentclass[runningheads]{llncs}

\usepackage{hyperref}

\usepackage[caption=false]{subfig}
\usepackage{graphicx}

\usepackage{paper1391}

\begin{document}

\title{Combining 3D~Model Contour Energy and~Keypoints for Object Tracking}
\titlerunning{3D~Model Contour Energy and~Keypoints for Object Tracking}

\author{
Bogdan Bugaev\inst{1,2}\orcidID{0000-0002-0994-1486}
\and\\
Anton Kryshchenko\inst{1,2}\orcidID{0000-0003-0088-1496}
\and\\
Roman Belov\inst{3}\orcidID{0000-0002-5481-6684}
}
\authorrunning{B. Bugaev, A. Kryshchenko, R. Belov}

\institute{
National Research University Higher School of Economics, St. Petersburg, Russia
\and
Saint Petersburg Academic University, St. Petersburg, Russia\\
\email{\{bogdan.bugaev,a.s.kryshchenko\}@gmail.com}
\and
KeenTools, St. Petersburg, Russia\\
\email{belovrv@gmail.com}
}

\maketitle

\begin{abstract}
We present a new combined approach for monocular model-based 3D~tracking.
A preliminary object pose is estimated by using a keypoint-based technique.
The pose is then refined by optimizing the contour energy function.
The energy determines the degree of correspondence between the contour of the
model projection and the image edges.
It is calculated based on both the intensity and orientation of the raw image
gradient.
For optimization, we propose a technique and search area constraints that allow
overcoming the local optima and taking into account information obtained
through keypoint-based pose estimation.
Owing to its combined nature, our method eliminates numerous issues of
keypoint-based and edge-based approaches.
We demonstrate the efficiency of our method by comparing it with
state-of-the-art methods on a public benchmark dataset that includes videos
with various lighting conditions, movement patterns, and speed.

\end{abstract}

\section{Introduction}

\begin{figure}[t]
\centering
\subfloat[]{
    \includegraphics{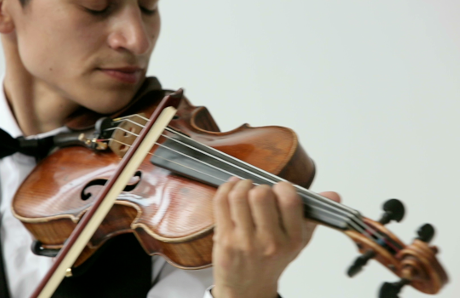}
    \label{fig:violin-frame}
}
\subfloat[]{
    \includegraphics{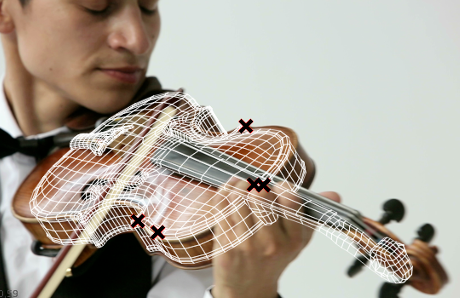}
    \label{fig:violin-tracked}
}
\subfloat[]{
    \includegraphics{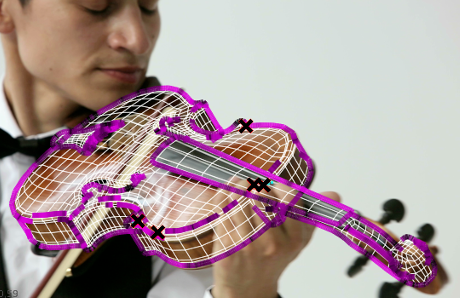}
    \label{fig:violin-optimized}
}
\caption{
    An example of a tracking algorithm applied to a single frame.
    \protect\subref{fig:violin-frame} A fragment of the processed frame.
    Despite the partial occlusions of the tracked object (\textit{violin}),
    most of its edges are visible.
    \protect\subref{fig:violin-tracked} The result of preliminary pose estimation.
    The model projection (\textit{white wireframe}) does not coincide with the
    object image in the frame because the position of the keypoints
    (\textit{black crosses}) used to determine its position was inaccurately
    calculated.
    \protect\subref{fig:violin-optimized} The object's model with optimized
    energy \ref{eqn:energy} of contours (\textit{purple lines}).
}\label{fig:violin}
\end{figure}

Monocular model-based 3D~tracking methods are an essential part of computer
vision. They are applied in a wide range of practical areas, from augmented
reality to visual effects in cinema. 3D~tracking implies iterative,
frame-by-frame estimation of an object's position and orientation relative
to the camera, with a given initial object pose. Fig.~\ref{fig:violin-frame}
shows a scene fragment typical for such a task. A number of characteristics
complicate tracking: the object is partially occluded, there are flecks and
reflections, and the background is cluttered.

In recent years, a great number of 3D~tracking methods have been developed.
These can be classified by the image characteristics and 3D~model features used
for pose detection. Many approaches
\cite{Lourakis2013,Vacchetti2004,ORBSLAM,Pauwels2013} are based on calculating
keypoints \cite{SIFT,ShiAndTomasi} on the image and corresponding points on the
3D~model.
Such methods make it possible to achieve high performance and robustness
against partial occlusions and cluttered background and are capable of
processing fast object movement.
At the same time, their use is limited for poorly textured objects because
keypoint calculation requires texture.

Nevertheless, objects tend to have one or other characteristic shape that lends
itself to being detected in an image.
Therefore, many methods use the information on the edges of the object's
3D~model projection---on its \term{contour} (for illustration, see
Fig.~\ref{fig:violin-optimized}).
As a rule, the contour of a 3D~object corresponds to the areas of an image that
are characterized by dramatic and unidirectional change in intensity---its
\term{edges}\cite{CANNY} on an image.
Methods
\cite{RAPID,Marchand2003,Choi2012,Marchand2006,Klein2006,SeoHinterstoisser2014,WangZhong2015,Damen2012,VacchettiEdges2004}
calculate the object pose from the correspondences between the points on the
3D~model contour and points on the image edges.
Some approaches are based on optimizing energy functions that determine the
degree of correspondence between the object projection in its current position
and its silhouette on an image
\cite{WangZhong2017,Marchand2001,PWP3D,Tjaden2017}.
Authors of \cite{Marchand2001} propose a tracking method using integral object
contour energy optimization; its value is the greater, the more precisely the
object contour fits the edges on the image.
The energy-based approaches risk detecting local energy function optima
that correspond to the wrong object poses.
In addition, edge-based methods have a drawback, namely the ambiguity
of symmetrical objects, which have identical contours in different
positions.

We present an approach that combines a keypoint-based method and integral edge
energy optimization.
A preliminary object pose is estimated using the Kanade--Lucas--Tomasi
(KLT)\cite{LucasAndKanade,TomasiAndKanade,ShiAndTomasi} tracker.
Then, we refine the object pose by optimizing the contour energy.
We modify the energy function described in \cite{Marchand2001} to take
into account the image edges' intensity as well as the directions of the model
contour and the edges.
We limit the search area for the optimal energy function value by using
the information obtained from the preliminary object pose estimation. This
allows for better convergence and makes it possible to partially overcome the
issue of symmetrical objects.
For optimization, we use the Basin-Hopping stochastic
algorithm\cite{BasinHopping} to avoid local optima.
In particular, we use an efficient Quasi-Newton method\cite{SLSQP} considering
the search area constraints directly.
Contour detection is a time-critical operation that is executed thousands of
times during the energy optimization.
We propose an algorithm that performs the most time-consuming computations as a
preliminary step.
This increases tracking performance.

In this paper, we concentrate on frame-to-frame object tracking that relies only
on the initial object pose and doesn't require any training using additional
data (such as predefined reference frames).

We demonstrate the efficiency of our approach as compared to state-of-the-art
3D~tracking methods in a series of experiments conducted on the OPT benchmark
dataset\cite{OPT}.
The dataset includes 552 test videos with ground truth object pose annotation.
The videos reproduce different lighting conditions and the main movement
patterns performed at a various speeds.
To compare the tracking efficiency, we use the metric proposed by the authors
of the OPT dataset.
Our test results demonstrate that, across the whole dataset, our method yielded
a value greater by $9.1\%$--$48.9\%$ with respect to the possible maximum than
other methods tested.
However, it should be noted that our method is not suitable for real-time
applications.

Further, we discuss the work related to the topic of this article in
Section~\ref{related-work}. Then, we provide a brief overview of the
mathematical notation in Section~\ref{input-data}. In Section~\ref{tracking},
we give a detailed description of the proposed tracking method. In
Section~\ref{evaluation}, we discuss the experimental results and provide a
comparison to other modern tracking methods. And, finally, in
Section~\ref{limitations}, we cover the limitations of our method as well as
our plans for its further improvement.

\section{Related Work}\label{related-work}

A detailed description of 3D~tracking methods can be found in
\cite{LepetitSurvey,MarchandSurvey}. In the present section, we shall discuss
solely the approaches based on the information on the contour of 3D~objects.

RAPID\cite{RAPID} was one of the first methods where the object pose was
estimated based on the correspondences between points on a 3D~model contour and
points on the edges of the image. To detect correspondences, they use
local 1D~search of the point with the largest edge intensity
along the perpendicular to the contour of the projected 3D~model.
In subsequent papers
\cite{Marchand2003,Choi2012,Marchand2006,Klein2006,SeoHinterstoisser2014}, a
number of improvements on the method were proposed; however, they were all
based on independent search for correspondences between points on the model contour and
points on the edges of the image. The main drawback of this approach lies in the fact
that the edge points of different objects can hardly be distinguished from each
other. This leads to 3D-2D correspondences containing a great number of
outliers, failing to preserve the object's contour, especially in case of
cluttered background, occlusions, or fast movement.

Other approaches introduce energy functions, where the value is the greater,
the greater the correspondence between the 3D~model projection and the image.
Therefore, the tracking goal can be achieved through the optimization of such energy. In
\cite{Marchand2001}, the authors propose using two variants of integrals along
the contour of the 3D~model projected onto the image gradient. One variant
takes into account only the absolute gradient value, while the other accounts
only for the direction of the gradient in points with sufficient absolute
gradient value. For energy optimization, the authors propose using coordinate
descent. For effective convergence, this method requires a very close
approximation to the sought-for optimum, which is calculated with the help of a
method similar to RAPID.

The approach described in the present article uses a similar energy function.
To improve convergence and overcome the issue of local optima, we use a global
optimization technique based on the Basin-Hopping algorithm
\cite{BasinHopping}. Methods in
\cite{Klein2006,WangZhong2017,Choi2012,ChoiFeaturesAndEdges} use a particle
filter to avoid local optima. In \cite{ChoiFeaturesAndEdges}, for particle
initialization, a keypoint-based approach is used, which leads to a more robust
algorithm and less ambiguity during tracking symmetrical objects. For
successful convergence in noise conditions and fast object movement, it becomes
necessary to use a great number of particles, which has a negative impact on
tracking performance. Unlike the particle filter, the Basin-Hopping algorithm
takes into account the information on the local optima that have already been
identified and makes it possible to use non-trivial termination criteria, thus
avoiding excessive calculations.

In addition to the information on the edge on the image, many methods also use
the information on color. In \cite{SeoHinterstoisser2014,WangZhong2015}, the
color distribution in the object and its background around the edge point is
used to eliminate false 3D-2D correspondences. Methods described in
\cite{PWP3D,Tjaden2017} optimize energy based on the color distribution in the
whole object and background. Such approaches are robust against partial
occlusions and cluttered scenes; however, they are sensitive to changes in lighting
and ambiguities arising from a similar coloring of the object and its
background.

\section{Input Data and Pose Parametrization}\label{input-data}

This section provides a brief overview of the mathematical notation used in the
present article.

The tracking algorithm accepts input data in the form of sequential
grayscale image frames $\Img_i \colon \ImgDom \to \Real$ (where
$\ImgDom \subset \Real^2$ is the image domain). Intensity of point
$\vect{u} \in \ImgDom$ in the frame $i$ equals $\Img_i(\vect{u})$. In cases
where the number of the frame is unimportant, the image is labeled as $\Img$.

The intrinsic parameters of the camera used to make the input frames are
assumed to be constant. They are given as the matrix
\begin{equation}\label{eqn:camintr}
    \CamIntr = \begin{bmatrix}
        f_x & 0   & c_x \\
        0   & f_y & c_y \\
        0   & 0   & 1
    \end{bmatrix}
    \text{.}
\end{equation}

3D~model $\Mesh$ describing the tracked object can be defined as $(\MeshV,
\MeshF)$, where $\MeshV \subset \Real^3$ is a finite set of model vertices, while
$\MeshF$ is the set of triplets of vertices defining model faces.

Object pose within frame $i$ has six degrees of freedom and is described as
\begin{equation}\label{eqn:modelpose}
    \Pose_i = \left[ \begin{array}{ccc|c}
          & \RotMat_i &   & \TrVec_i \\
        \hline
        0 & 0         & 0 & 1
    \end{array} \right] \in \SE
    \,\text{,}
\end{equation}
where $\RotMat_i \in \SO$ defines the orientation of the model and
$\TrVec_i \in \Real^3$ defines its translation.

The projection $\vect{u} \in \Real^2$ of a point on the model surface
$\vect{x} \in \Real^3$ is described by a standard camera model
\begin{equation}\label{eqn:proj}
    \homv{u} = \CamIntr \cdot [\RotMat_i | \TrVec_i] \cdot \homv{x}
    \,\text{,}
\end{equation}
where $\homv{u}$ and $\homv{x}$ are vectors $\vect{u}$ and $\vect{x}$
in homogeneous coordinates. The function performing the projection
$\vect{x} \mapsto \vect{u}$ in the pose $\Pose_i$ shall be designated as
$\Projection{\Pose_i}$.

\section{3D Tracking by Combining Keypoints and Contour Energy}\label{tracking}

We solve the tracking task in two steps. During the first step, the
Kanade--Lucas--Tomasi (KLT)
tracker\cite{LucasAndKanade,TomasiAndKanade,ShiAndTomasi}
is applied to estimate object pose.
During the second step, object pose is refined by optimizing the objective
function, i.e.\ the contour energy optimization.

The present section provides a detailed description of the proposed tracking
method. First, we give a brief overview of the initial object pose estimation
algorithm. Further, we concentrate on the method for pose refinement. First of
all, we give a detailed description of the contour energy. Then, we
discuss its optimization: we provide an overview of the global optimization
method and the local optimization method that it utilizes. Then, we propose a
step-by-step procedure to refine the object pose by using the method described.
After that, we discuss bound constraints estimation for the energy optimum
search area. And, finally, we describe the object contour detection algorithm.

\subsection{Initial Object Pose Estimation}\label{tracking:estimation}

We estimate the initial object pose with the help of a wide-known KLT tracker.
On the frames where the object pose is known, we identify 2D~keypoints and
determine corresponding 3D~points on the surface of the model. Keypoint
movement is tracked with the help of optical flow calculation. On the image,
the known 2D-3D correspondences are used to estimate the object pose by solving
the \PnP{} problem\cite{LepetitSurvey} while using RANSAC\cite{RANSAC} to
eliminate outliers.

When after a sufficiently great number of iterations RANSAC fails to find a
solution with an acceptable percentage of outliers, or when the number of
points tracked is very small, we estimate object pose in frame $i$ by
extrapolation based on poses in frames  $i - 1$ and $i - 2$.

\subsection{Contour Energy}

We view pose optimization as the matching of model contour with object edges
in the image. Model contours are understood as two types of lines.
Firstly, it is the outer contour of the projected model.
Secondly, it is the projections of visible sharp model edges. Sharp
model edges are the edges where adjacent faces meet at an angle no greater than
the pre-selected one $\MinAngle$. Fig.~\ref{fig:violin-optimized} provides an
example of such matching.

To further ideas described in \cite{Marchand2001}, we suggest the following
energy function for a quantitative expression of matching quality:
\newcommand{\ImgGradAtPoint}{\nabla \Img(\CoordAt(s))}
\newcommand{\NormalAtPoint}{\NormalAt(s)}
\newcommand{\EnergyDot}{\abs{\ImgGradAtPoint \cdot \NormalAtPoint}}
\begin{equation}\label{eqn:energy}
    \EnergyAt = \frac{\mathlarger{
        \int_{\ContourAt} \EnergyDot \, ds
    }}{\mathlarger{
        \int_{\ContourAt} ds
    }}
    \,\text{,}
\end{equation}
where $\ContourAt$ are model contour lines, $\CoordAt$ is the function
returning the contour point coordinate in the image, $\NormalAt$ is the
function returning the normal unit vector to contour.

The energy is an integral characteristic of the contour (numerator) normalized
along the length of the contour (denominator). The division by the length of
the contour is done to avoid the case where the long contour is preferred to
the shorter one.

Let us consider the numerator expression under integral sign:
\begin{equation}
    \EnergyDot
    \,\text{.}
\end{equation}
The image gradient $\ImgGradAtPoint$ shows the direction and strength of
intensity change in a point. If there is an edge, gradient is perpendicular to
it. The unit vector $\NormalAtPoint$ is perpendicular to the model contour.
The absolute value of their scalar product is the greater, the more visually
significant the edge in the image is (i.e.\ the greater the gradient magnitude)
and the higher the correspondence of its direction in the current point and the
direction of the model contour.

Therefore, the value $\EnergyAt$ is the greater, the greater the correspondence of
the model contour in pose $\Pose$ and the edges in the image and the more
visually significant those edges (for example, see Fig.~\ref{fig:violin-optimized}). Given
that the object edges are sufficiently visible, the energy, in most cases,
will be maximal in the sought-for pose. Therefore, the optimal object pose in frame
$i$ can be found as
\begin{equation}
    \Pose_i = \argmax_{\Pose} \EnergyAt
    \,\text{.}
\end{equation}

Due to integral nature of the energy function \ref{eqn:energy}, it is
sufficiently robust against occlusions.
Its disadvantage lies in the fact that, potentially, cases
where the wrong object pose will have greater energy than its true pose are
possible. However, practical experience shows that such cases are quite rare.
In addition, ambiguities in detecting the pose of objects of a symmetrical
(e.g., cylindrical) shape may be observed.

To implement the evaluation of contour energy, it is necessary to perform the
discretization of expression \ref{eqn:energy}:
\begin{equation}\label{eqn:energysum}
    \EnergySumAt =
        \frac{1}{\abs{\ContourPointsAt}}
        \sum_{s \in \ContourPointsAt} \EnergyDot
    \,\text{,}
\end{equation}
where $\ContourPointsAt$ is the finite set of points uniformly distributed
along the contour lines.
\let\EnergyDot\undefined
\let\NormalAtPoint\undefined
\let\ImgGradAtPoint\undefined

A detailed description of contour line detection is described in
Section~\ref{tracking:contours}.

\subsection{Energy Optimization Method}\label{tracking:optmethods}

\begin{figure}[t]
\includegraphics{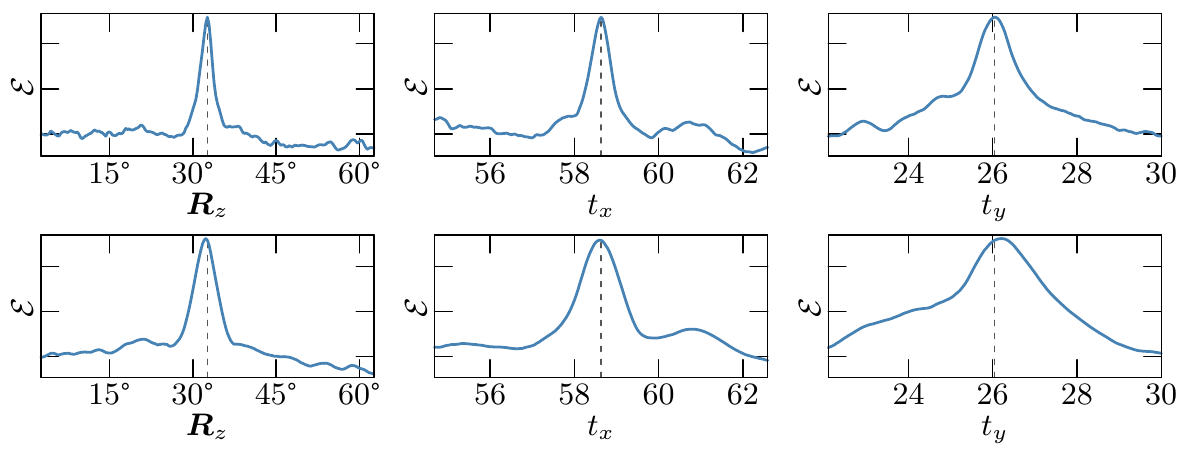}
\caption{
    Examples of contour energy \ref{eqn:energy} near an optimum. In the top row
    the energy was calculated from the original image; in the bottom row it was
    calculated from an image blurred by convolution with a Gaussian kernel.
    Label $\RotMat_z$ denotes the dependence on the rotation around axis $z$
    (the other degrees of freedom being fixed), $\TrVec_x$ and $\TrVec_y$
    denote the dependence on translation along axes $x$ and $y$ respectively.
    The areas demonstrated here correspond $60^\circ$ for the rotation and
    approximately $0.2$ of object size.
}
\label{fig:slices}
\end{figure}

In most cases, contour energy \ref{eqn:energy} has a notable global
optimum in the area of the sought-for object pose and, at the same time, shows
a plenty of local optima at a certain distance from it (see Fig.~\ref{fig:slices}).

In many cases, the first approximation obtained during object pose estimation
is not in the concave region near the sought-for optimum. On the other hand, we may
assume that such first approximation will turn out to be good enough;
therefore, we propose to limit the search area and then apply an optimization
method capable of overcoming the local optima. A detailed description of search
area bounds estimation is given in Section~\ref{tracking:bounds}.

For optimization, the version of the Basin-Hopping stochastic algorithm
described in \cite{BasinHopping} was selected. Basin-Hopping is an iterative
algorithm. At each step, a random hop within the search area is made; after
that, local optimization is performed and, then, the obtained local optimum is
either accepted or rejected based on the Metropolis criterion\cite{Metropolis}.
The algorithm stops once the maximum number of iterations
has been reached or if several previous steps did not serve to improve the
optimum and the minimal number of iterations has been performed.

For local optimization, the SLSQP algorithm was selected\cite{SLSQP}. It
combines the option of limiting the search area and the efficiency of
Quasi-Newton methods.

The energy gradient is computed numerically.

To improve convergence, we blur the frame by convolution $\GaussS \ast \Img$
with the Gaussian kernel $\GaussS$ before optimization.
Blurring effectively suppresses noise and, along with it, high-frequency
details.
It has a positive impact on the smoothness of the energy function, but may
cause a slight displacement of the sought-for optimum
(see Fig.~\ref{fig:slices}).
We eliminate this displacement via adding a final optimization step using the
original image.

\subsection{Search Area Bounds}\label{tracking:bounds}

During initial pose estimation (Section~\ref{tracking:estimation}), the object
pose can be obtained either with the help of the KLT tracker or through
extrapolation based on the previous poses. The former differs from the latter
in that there is extra data present; therefore, we shall consider them
separately.

In case the object pose could be obtained only through extrapolation, we
propose selecting the maximal deviations from the estimated pose based on the
assumption on the degree of object movement in consecutive frames. In our
experiments, we have limited the rotation around each Euler angle to
$\pm 30^\circ$, the translation along the camera axis to $\pm 0.2$ of the
object size, and the translation along the other axes to $\pm 0.1$ of the
object size.

In case of success, the KLT tracker estimates the object pose by solving the
\PnP{} problem on a set of 2D and 3D~point correspondences
$\{(\vect{u}_1, \vect{x}_1), \dots, (\vect{u}_{\InlierCount}, \vect{x}_{\InlierCount})\}$
by minimizing the average reprojection error
$\Approx{\Pose} = \argmin_{\Pose} \AvgReprojErr(\Pose)$,
where
\begin{equation}
    \AvgReprojErr(\Pose) =
        \frac{1}{\InlierCount} \sum_{j = 1}^{\InlierCount}{
            \norm{\vect{u}_j - \Projection{\Pose}(\vect{x}_j)}
        }
    \,\text{.}
\end{equation}

The average reprojection error $\AvgReprojErr(\Pose)$ can be understood as the
measure of consistency of object pose $\Pose$ with the position of keypoints
used to reach the solution: the smaller the error, the greater the consistency.
Due to errors arising during keypoint tracking, the object pose that is most
consistent with them may be different from its true pose, but it will be in its
near neighborhood.

We propose selecting search area bounds in such a way that, when approaching
them, $\AvgReprojErr$ does not increase by a value greater than pre-selected
one $\ErrMax$, i.e.\ the consistency with keypoints does not deteriorate below
a given threshold:
\begin{equation}
    \AvgReprojErr(\Pose) - \AvgReprojErr(\Approx{\Pose}) \leq \ErrMax
    \,\text{.}
\end{equation}
The optimization methods proposed in Section~\ref{tracking:optmethods} make it
possible to set such non-linear constraints on the search area directly.

The size of the search area usually is quite natural in practice.
For example, it is often the case that a very
noticeable change of object pose in certain directions leads to a relatively
small change in average reprojection error. This mostly concerns movement along
the camera axes. This can also happen when keypoints are not evenly distributed
throughout the object and cover it only partially. Along with errors in
keypoint position, this normally results in noticeably inaccurate object pose
estimation; for example, see Fig.~\ref{fig:violin-tracked}. Setting the bounds
accounting for average reprojection error makes it possible to obtain a broad
enough search area and then use energy optimization to find a rather precise
pose, as shown in Fig.~\ref{fig:violin-optimized}.

\subsection{Model Contour and Sharp Edges Detection}\label{tracking:contours}

Visible contour and sharp edges detection algorithms can be grouped into two
categories. The first type of algorithms is based on model rendering.
They are precise and allow for correct processing of
self-occlusions. However, rendering requires time-consuming computing.
The second type of algorithms is based on the analysis of the model
itself: its edges and the spatial relationship between the adjacent faces. They
are less complex in terms of computing, but they fail to account for
self-occlusion.

We propose a combined approach. Prior to tracking, we perform the most
time-consuming calculations and gather data on the visibility of model faces
from various points of view. After that, while estimating the contours during tracking,
in a single run, we identify sharp and contour edges and process self-occlusion
with the help of this data.

When calculating energy \ref{eqn:energy}, the object pose $\Pose$ for which the
contours need to be estimated is known. For this purpose, all model $\Mesh$
edges formed by two faces are reviewed. Out of these, the ones lying on the
contour or formed by faces meeting at an acute angle are selected.

\newcommand{\MeshVxP}{\vect{p}}
\newcommand{\MeshVxQ}{\vect{q}}
\newcommand{\MeshVxA}{\vect{v}_1}
\newcommand{\MeshVxB}{\vect{v}_2}
\newcommand{\FaceA}{(\MeshVxA, \MeshVxQ, \MeshVxP)}
\newcommand{\FaceB}{(\MeshVxB, \MeshVxP, \MeshVxQ)}

\begin{figure}[t]
\centering
\subfloat[]{
    \includegraphics{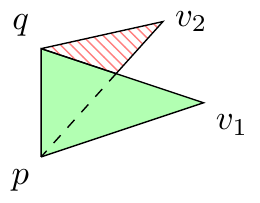}
    \label{fig:contour-edge}
}
\hspace{0.02\textwidth}
\subfloat[]{
    \includegraphics{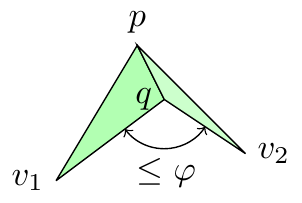}
    \label{fig:sharp-edge}
}
\subfloat[]{
    \includegraphics{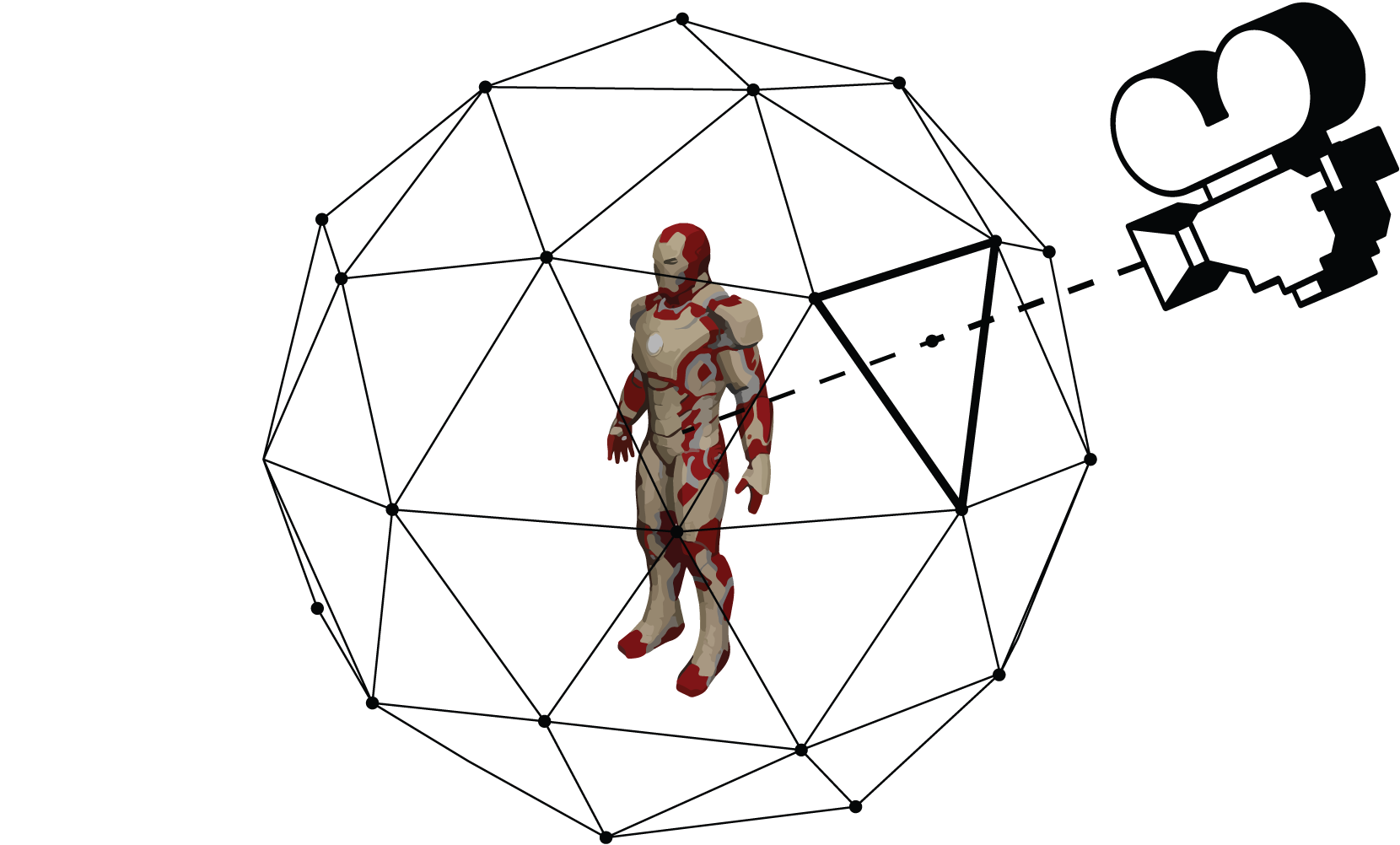}
    \label{fig:icosphere}
}
\caption{
    Edge $\MeshVxP \MeshVxQ$ type recognition.
    \protect\subref{fig:contour-edge} Contour edge is formed by front face
    $\FaceA$ (marked by \textit{green color}) and back face $\FaceB$
    (marked by \textit{pink hatch}).
    \protect\subref{fig:sharp-edge} A sharp edge is formed by two front faces
    angled at under $\MinAngle$.
    \protect\subref{fig:icosphere} Our contour detection algorithm rejects the
    edge if any adjacent front face is invisible from all of three the nearest
    preprocessed viewpoint directions.
}\label{fig:edges}
\end{figure}

Let us consider edge $\MeshVxP \MeshVxQ$ and its adjacent faces $\FaceA$ and
$\FaceB$. The edge is considered sharp if both of its faces are turned with
their outer surface towards the camera and the angle between them is no
greater then $\MinAngle$ (see Fig.~\ref{fig:sharp-edge}). The edge is
considered to be lying on the contour if one of its faces is turned towards
the camera and the other is not (see Fig.~\ref{fig:contour-edge}).
It is obvious that some of the edges identified in this manner will be
invisible in case of self-occlusions. To eliminate a major part of invisible
edges, it is sufficient to know which of the front faces are invisible for the
current point of view.

\let\MeshVxP\undefined
\let\MeshVxQ\undefined
\let\MeshVxA\undefined
\let\MeshVxB\undefined
\let\FaceA\undefined
\let\FaceB\undefined

Data on model face visibility is collected prior to tracking. For this purpose,
the model is rendered from several viewpoint directions with orthographic
projection camera.
To reach uniform distribution of these directions, we use vertices of
an icosphere (recursively divided icosahedron) surrounding the model.
Given an object pose,
we determine three preprocessed viewpoint directions that are close to
the current direction and form a face of the icosphere as shown in
Fig.~\ref{fig:icosphere}.
The faces invisible from this three directions will be likely invisible in the
current pose.
Knowing such faces, we can exclude adjacent contour
and sharp edges.
More formal description of self-occlusion processing can be
found in supplementary material.

Having detected contour and sharp edges and having processed self-occlusion, it
is easy to project the edges onto the frame and select points for numerical
calculation with the formula \ref{eqn:energysum}.

An obvious drawback of the described approach is that it is not perfectly
precise in self-occlusion processing, which can lead to a certain percentage
of wrongly detected contours. In true model pose, they are unlikely to correspond to
the edges on the image, but may potentially occur on some of the edges in the
wrong pose. However, in most cases, a small amount of such false contours does
not lead to wrong pose estimation.

The advantage of the described algorithm is that the calculations performed
during tracking are relatively simple and less computationally complex compared
to algorithms that require rendering.

\section{Evaluation}\label{evaluation}

To prove the efficiency of the method described in the present article, we have
tested it on the OPT dataset\cite{OPT} and compared our results with those
obtained from state-of-the-art tracking approaches. The test dataset includes RGB
video recordings of tracked objects, their true pose, and 3D~models. The videos
are $1920\times1080$ and have been recorded with the help of a programmable
robotic arm under various lighting conditions. Object movement patterns are
diverse and many in number and the velocity of movement also varies. OPT
contains six models of various geometric complexity. For each of these, 92 test
videos of varying duration ($23$--$600$ frames) have been made, the total
number of frames being $79968$. The diverse test data covers most of the motion
scenarios.

Our method has been implemented in C++ and is part of a software product
intended for 3D~tracking for film post-production. All experiments were
conducted on a computer with an Intel i7-6700 $3.4$~GHz processor and $32$~GB
of RAM. Details of our method settings are given in supplementary material.

In the present section, we first describe the approach for results evaluation
used to compare our method with other tracking methods. Further, we show the
efficiency of object pose optimization based on contour energy. Then, we
compare the results from our approach to those obtained by other modern
tracking methods. To conclude, we discuss the advantages and drawbacks of our
method as well as potential ways for its improvement.

\subsection{Evaluation Metric}

\newcommand{\TruePose}{\hat{\Pose}}
\newcommand{\PoseErr}{\delta}
\newcommand{\Diameter}{d}

Given the known true object pose $\TruePose_i$ in frame $i$, we calculate the
estimated pose $\Pose_i$ error as
\begin{equation}
    \PoseErr_i = \avg_{\homv{x} \in \MeshV} \norm{\TruePose_i \homv{x} - \Pose_i \homv{x}}
    \text{,}
\end{equation}
where $\MeshV$ is the set of 3D~model vertices. We consider the object pose
within the frame successfully detected if $\delta_i$ is less than $k d$, where
$d$ is the diameter of the 3D~model and $k$ is a given error coefficient.

To compare the efficiency of different methods, we create a curve where each
point is defined as the percentage of frames where object pose with respect to
varying $k$ was successfully determined. The more efficient method of object
pose tracking corresponds with the greater value of AUC (area under curve). In
our experiments, $k$ varies from $0$ to $0.2$; therefore AUC varies from
$0$ to $20$.

\subsection{Effectiveness of the Contour Energy Optimization}

\begin{figure}[t]
\centering
\includegraphics{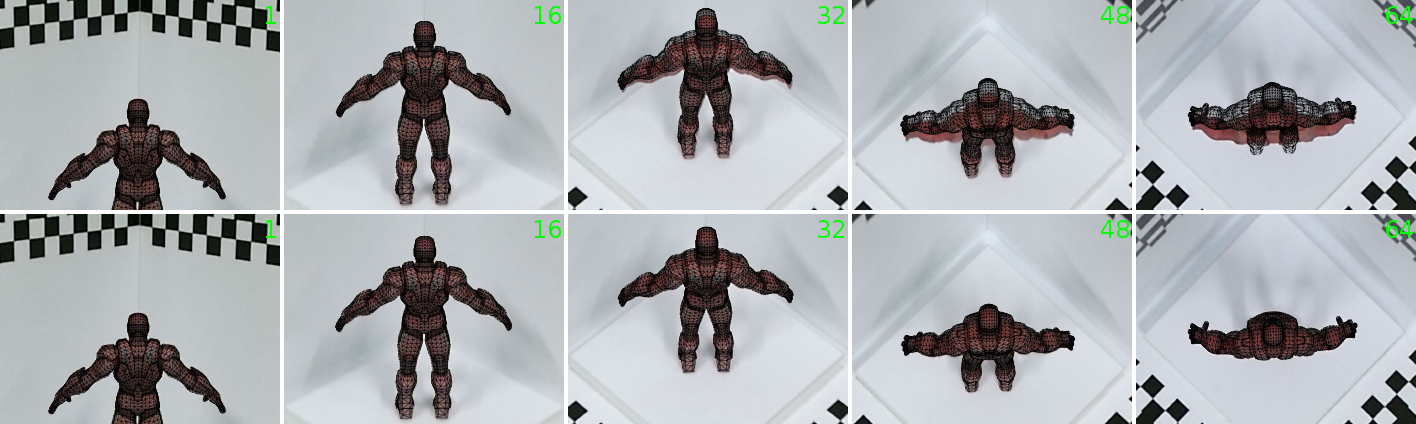}
\caption{
    An example of tracking efficiency increase due to contour energy
    optimization.
    Upper and lower rows show video frame fragments (the frame numbers are in
    the top right corners) with the 3D~model (\textit{black wireframe})
    projected in the pose estimated with the help of pure KLT tracker and with
    the help of our method, respectively.
}\label{fig:ironman}
\end{figure}

Let us show how the contour energy optimization improves the results of pure
Kanade--Lucas--Tomasi tracker using an example; see Fig.~\ref{fig:ironman}.
When the object poses obtained with the help of KLT tracker are not refined
through the contour energy optimization, the object gradually descends onto the
background. In the final frames the discrepancy with the actual object pose
becomes significant. This behavior is due to frame-by-frame error accumulation.
The refinement step using the contour energy optimization eliminates the errors
and results in keeping the object close to its actual pose in all frames.

Experiments conducted on the OPT dataset confirm a significant increase in
tracking efficiency due to contour energy optimization. Thus, in all tests, the
AUC value increased by $18\%$, while in the group of tests with maximum
movement velocity the increase was $27\%$.

\subsection{Comparison with State-of-the-Art Methods}

We have compared our approach with the following state-of-the-art methods:
GOS\cite{WangZhong2015}, PWP3D\cite{PWP3D} and ORB-SLAM2\cite{ORBSLAM}.
GOS is an edge-based method improved the approach from \cite{SeoHinterstoisser2014}.
It takes into account color distribution around the edges and contour coherence.
PWP3D tracks object by segmenting the frame into object and
background and using color distribution statistics.
ORB-SLAM2 is state-of-the-art simultaneous localization and mapping approach
based on keypoints detection. The authors of \cite{OPT} proposed modifications
that allowed applying this method to 3D~model tracking.
Tracking results for PWP3D and ORB-SLAM2 applied to the OPT dataset are
cited from \cite{OPT}. Tracking results for GOS were received from
testing the open source implementation. During testing, all methods were
initialized with the true object pose in the first frame of each video.

\begin{figure}
\includegraphics{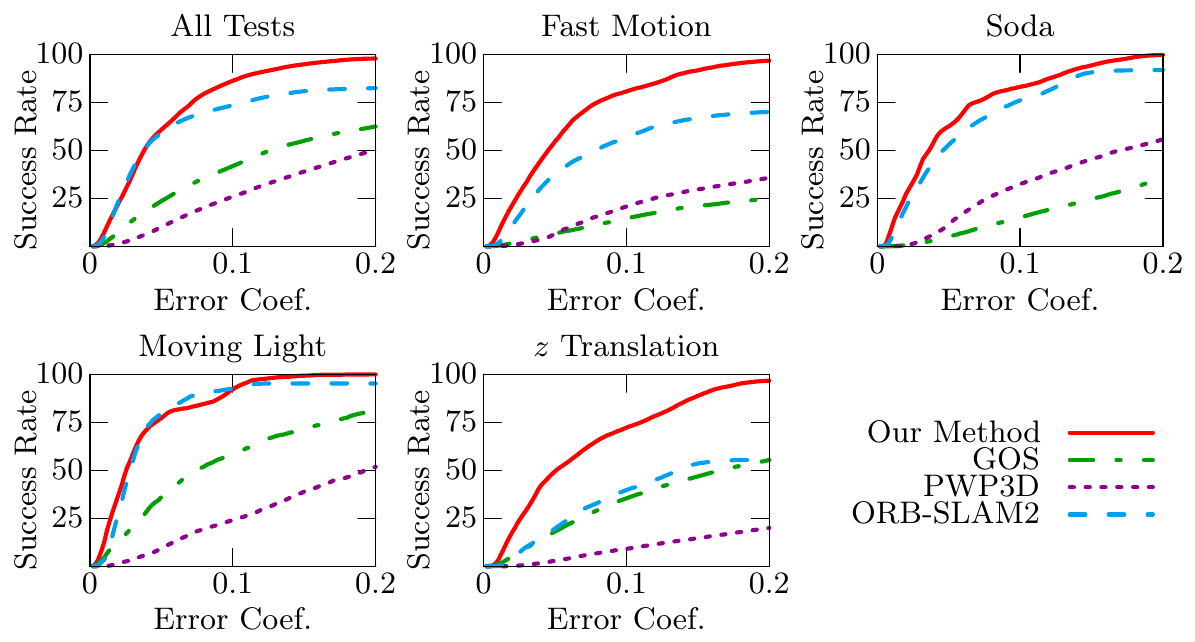}
\caption{
    Comparison of method efficiency on various test groups in OPT
}
\label{fig:resAll}
\end{figure}

Fig.~\ref{fig:resAll} demonstrates the results from some of the test groups in
OPT. Tables~\ref{tab:results-by-models} and~\ref{tab:results-by-groups} contain the
complete detailed testing results. Overall, all tests show a noticeable
disadvantage of other methods as compared to ours: GOS by 35.6\%, PWP3D by 48.9\%, ORB-SLAM2
by 9.1\% (all values have been calculated with respect to the maximum possible
value).

Table~\ref{tab:perf-by-methods} contains average frame processing times
of tested methods. ORB-SLAM2 and PWP3D are positioned as real-time methods
and they show the best run time performance.
At the same time, our method and GOS are not suitable for real-time
applications and our method is in $3.4$ times slower than GOS on average.
In fact, run time of our method significantly depends on the size of object model
as shown in Table~\ref{tab:perf-by-models}.

More detailed results of experiments can be found in the supplementary
material.

\begin{table}
\begin{center}
    \caption{Comparison of AUC in tests with different objects}
    \label{tab:results-by-models}
    \begin{tabular}{p{0.2\textwidth} p{0.11\textwidth} p{0.1\textwidth}
    p{0.1\textwidth} p{0.1\textwidth} p{0.11\textwidth} p{0.1\textwidth}
    p{0.1\textwidth}}
    \hline
    Approach       & All tests   & Bike        & Chest       & House       & Ironman     & Jet         & Soda        \\
    \hline
    Our method     & {\bf 14.79} & {\bf 12.55} & 14.97       & 14.48       & {\bf 14.71} & {\bf 17.17} & {\bf 14.85} \\
    GOS            & 7.68        & 3.38        & 11.28       & 8.49        & 9.91        & 9.89        & 3.12        \\
    PWP3D          & 5.01        & 5.36        & 5.55        & 3.58        & 3.92        & 5.81        & 5.87        \\
    ORB-SLAM2      & 12.97       & 10.41       & {\bf 15.53} & {\bf 17.28} & 11.35       & 9.93        & 13.44       \\
    \hline
    \end{tabular}
\end{center}
\end{table}

\begin{table}
\begin{center}
    \caption{Comparison of AUC in tests with different tracking conditions}
    \label{tab:results-by-groups}
    \begin{tabular}{p{0.18\textwidth} p{0.09\textwidth}
    p{0.09\textwidth} p{0.11\textwidth} p{0.09\textwidth} p{0.11\textwidth}
    p{0.13\textwidth} p{0.13\textwidth}}
    \hline
    Approach & Flash light        & Free motion         & Rotation
             & Fast motion        & Moving light        & {$x$-$y$ \newline translation}
             & {$z$ \newline translation} \\
    \hline
    Our method     & 13.96       & {\bf 13.91} & {\bf 15.51} & {\bf 14.06} & {\bf 16.36} & {\bf 14.76} & {\bf 13.08} \\
    GOS            & 10.92       & 0.95        & 9.01        & 2.75        & 10.51       & 6.74        & 6.45        \\
    PWP3D          & 5.08        & 2.89        & 5.89        & 3.76        & 4.91        & 8.73        & 1.83        \\
    ORB-SLAM2      & {\bf 15.91} & 9.10        & 15.24       & 9.91        & 15.99       & 13.30       & 7.07        \\
    \hline
    \end{tabular}
\end{center}
\end{table}

\begin{table}
\begin{center}
    \caption{Comparison of average frame processing time (ms)}
    \label{tab:perf-by-methods}
    \begin{tabular}{p{0.2\textwidth} p{0.15\textwidth} p{0.2\textwidth} p{0.2\textwidth}}
    \hline
    Our method  & GOS  & PWP3D  & ORB-SLAM2 \\
    \hline
    683         & 201  & 66     & 67 \\
    \hline
    \end{tabular}
\end{center}
\end{table}

\begin{table}
\begin{center}
    \caption{Dependency of our method average frame processing time on object size}
    \label{tab:perf-by-models}
    \begin{tabular}{p{0.15\textwidth} p{0.12\textwidth} p{0.12\textwidth} p{0.12\textwidth}
    p{0.12\textwidth} p{0.12\textwidth} p{0.12\textwidth}}
    \hline
    Value             & Bike     & Chest    & House   & Ironman   & Jet    & Soda \\
    \hline
    time (ms)         & 1097     & 358      & 550     & 427       & 1301   & 364 \\
    \abs{\MeshF}      & 156950   & 28648    & 2594    & 11496     & 176260 & 6788 \\
    \hline
    \end{tabular}
\end{center}
\end{table}

\subsection{Discussion and Future Work}\label{limitations}

Testing results provided in
Tables~\ref{tab:results-by-models},~\ref{tab:results-by-groups}
and in Fig.~\ref{fig:resAll} show that our
method demonstrated good results under various tracking conditions.

Tests under moving light and flashing light show high performance in spite of
the fact that the KLT tracker lacks robustness against dramatic lighting
changes. It is also worth noting that our method significantly better processes
object movement along the camera axis (‘$z$ translation’ tests) in comparison with
other methods.
Edge-based methods are very sensitive to motion blur. Nevertheless, the results
of ‘Fast Motion’ tests demonstrate that our method handles motion blur and fast
movement more efficiently than other methods.

Symmetrical objects in different poses may have identical contours,
which leads to ambiguity during contour-based object pose estimation. By
limiting the pose search area, we mostly overcome this issue. Test results
from a symmetrical object, Soda,
confirm this finding. Our method successfully tracks this object while
the GOS---another edge-based method---shows low efficiency.

Disadvantage of our method is the negative impact that lack of model
accuracy has on tracking quality.
This is presented in Table~\ref{tab:results-by-models}: tracking results for
the Bike object are noticeably lower than those for other objects.
Our method is also not devoid of the
drawback typical for most edge-based approaches---jitter. To improve the
efficiency of the method in the above-mentioned cases, in the future,
we are planning to use edges on the inner texture of the object.

A major limitation of our method is low run time performance
(see Table~\ref{tab:perf-by-methods}).
Also, frame processing time significantly depends
on object model size as shown in Table~\ref{tab:perf-by-models}.
This is due to the following fact: the most time-consuming part of our method is
contour energy optimization, where model contour identification is the operation
repeated most frequently. These calculations could be accelerated using GPU.

\section{Conclusions}

The present article introduced the method for model-based 3D~tracking based on
combination of model contour energy and keypoints. Kanade--Lucas--Tomasi
tracker is used for preliminary pose estimation. Then the pose is refined by
the contour energy optimization using the Basin-Hopping stochastic algorithm.
To improve optimization, we set search area constraints based on keypoints
average reprojection error. Such constraints fix well-tracked parts of the
object while allows movement of parts which are failed to be correctly
positioned by KLT tracker. The results of experiments on a challenging
benchmark dataset show that combining edge-based and keypoint-based approaches
can diminish the typical disadvantages of both methods. Contour energy optimization
effectively struggle with motion blur and poorly textured surfaces, while
keypoints help to correctly process symmetrical objects.
We demonstrated the efficiency of our approach by testing it against
state-of-the-art methods on a public benchmark dataset.

\end{document}


\title{Combining 3D~Model Contour Energy and~Keypoints for Object Tracking}
\subtitle{Supplementary Material}
\titlerunning{Supplementary Material}

\author{
Bogdan Bugaev
\and
Anton Kryshchenko
\and
Roman Belov
}
\authorrunning{B. Bugaev, A. Kryshchenko, R. Belov}

\institute{}

\maketitle

\section{Summary}\label{summary}

This document contains full results of performed tests and some details that
are not essential to the understanding of the main paper.
In Section~\ref{settings} we describe general settings of tested implementation
of our method. Section~\ref{evaluation} contains full evaluation results.
In Section~\ref{contour} we describe details of 3D~model contour detection.

\section{Implementation Settings}\label{settings}

\paragraph{KLT Tracker}

We consider that KLT tracker fails if there is less then 8 keypoints or
if inlier rate is less then $0.3$.

\paragraph{Contour Detection}

A 3D~model edge is considered as sharp if it is formed by two faces angled at
under $\MinAngle = \ang{45}$. To generate face visibility data, we use
2562 viewpoint directions.

\paragraph{Search Area Bounds}

If the object pose is successfully estimated using KLT tracker, maximum
reprojection error threshold $\ErrMax = 2.5$ is used for selecting search area
bounds. If not, we limit the rotation around each Euler angle to $\pm\ang{30}$,
the translation along the camera axis to $\pm 0.2$ of the object size, and the
translation along the other axes to $\pm 0.1$ of the object size.

\paragraph{Energy Optimization}

The Basin-Hopping algorithm stops once the maximum number of iterations has
been reached or if several previous steps did not serve to improve the optimum
and the minimal number of iterations has been performed. To set the termination
criterion, we calculate a coefficient
\newcommand{\Coef}{s}
\[
    \Coef = \max \left\{ 1, \frac{25000}{\abs{\MeshV} + \abs{\MeshF}} \right\}
\]
at first. Then we set the minimum number of hops to
$\min\{10 \Coef, 100\}$,
the maximum number of effectless hops to
$\min\{5 \Coef, 30\}$ and
the maximum number of hops to
$\min\{30 \Coef, 200\}$.
\let\Coef\undefined
It allows to perform a bit more iterations if the 3D~model size is not too big.
Before optimization, we blur the frame by convolution with Gaussian blur kernel
$\GaussS$, standard deviation is $\SD = 1.1$.
Then we run optimization again on the original frame using fixed number of hops
that is equal to 5.

\section{Evaluation}\label{evaluation}

\begin{figure}[t]
\centering
\subfloat[]{
    \includegraphics{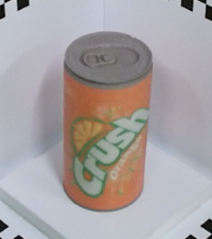}
    \label{fig:objects-soda}
}
\subfloat[]{
    \includegraphics{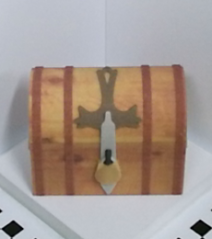}
    \label{fig:objects-chest}
}
\subfloat[]{
    \includegraphics{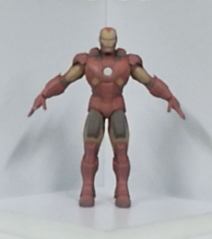}
    \label{fig:objects-ironman}
}
\subfloat[]{
    \includegraphics{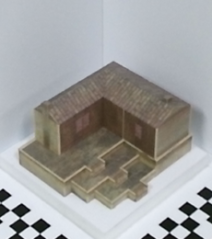}
    \label{fig:objects-house}
}
\subfloat[]{
    \includegraphics{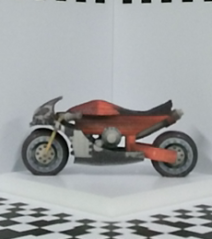}
    \label{fig:objects-bike}
}
\subfloat[]{
    \includegraphics{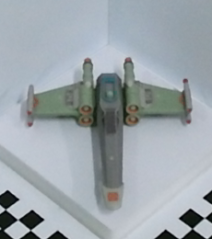}
    \label{fig:objects-jet}
}
\caption{
    Tracked objects:
    \psref{objects-soda} soda,
    \psref{objects-chest} chest,
    \psref{objects-ironman} ironman,
    \psref{objects-house} house,
    \psref{objects-bike} bike,
    \psref{objects-jet} jet
}\label{fig:objects}
\end{figure}

We tested our method using the OPT benchmark dataset\cite{OPT}.
It contains videos of 3D~objects with simple (\term{soda}, \term{chest}),
normal (\term{ironman}, \term{house}), and complex (\term{bike}, \term{jet})
geometry (see Fig.~\ref{fig:objects}).
Also the OPT includes the following test groups:
\begin{itemize}
\item \term{$x$-$y$ translation}. An object moves along a circle parallel to
the camera sensor plane with motion blur in all directions.
\item \term{$z$ translation} (or \term{zoom}). An object moves forward first
and then backward along an axis perpendicular to the camera sensor plane.
\item \term{In-plane rotation}. An object rotates along an axis perpendicular
to the camera sensor plane.
\item \term{Out-of-plane rotation}. An object rotates along an axis parallel to
the camera sensor plane.
\item \term{Flash light}. The light source is turned on and off repeatedly, and
the object moves slightly.
\item \term{Moving light}. The light source moves and results in illumination
variations while the object moves slightly.
\item \term{Free motion}. An object moves in arbitrary directions.
\end{itemize}
In \term{translation} and \term{rotation} tests the objects move at five
different speed levels, from slow to fast.

\subsection{Tracking Efficiency}

\begin{figure}
\includegraphics{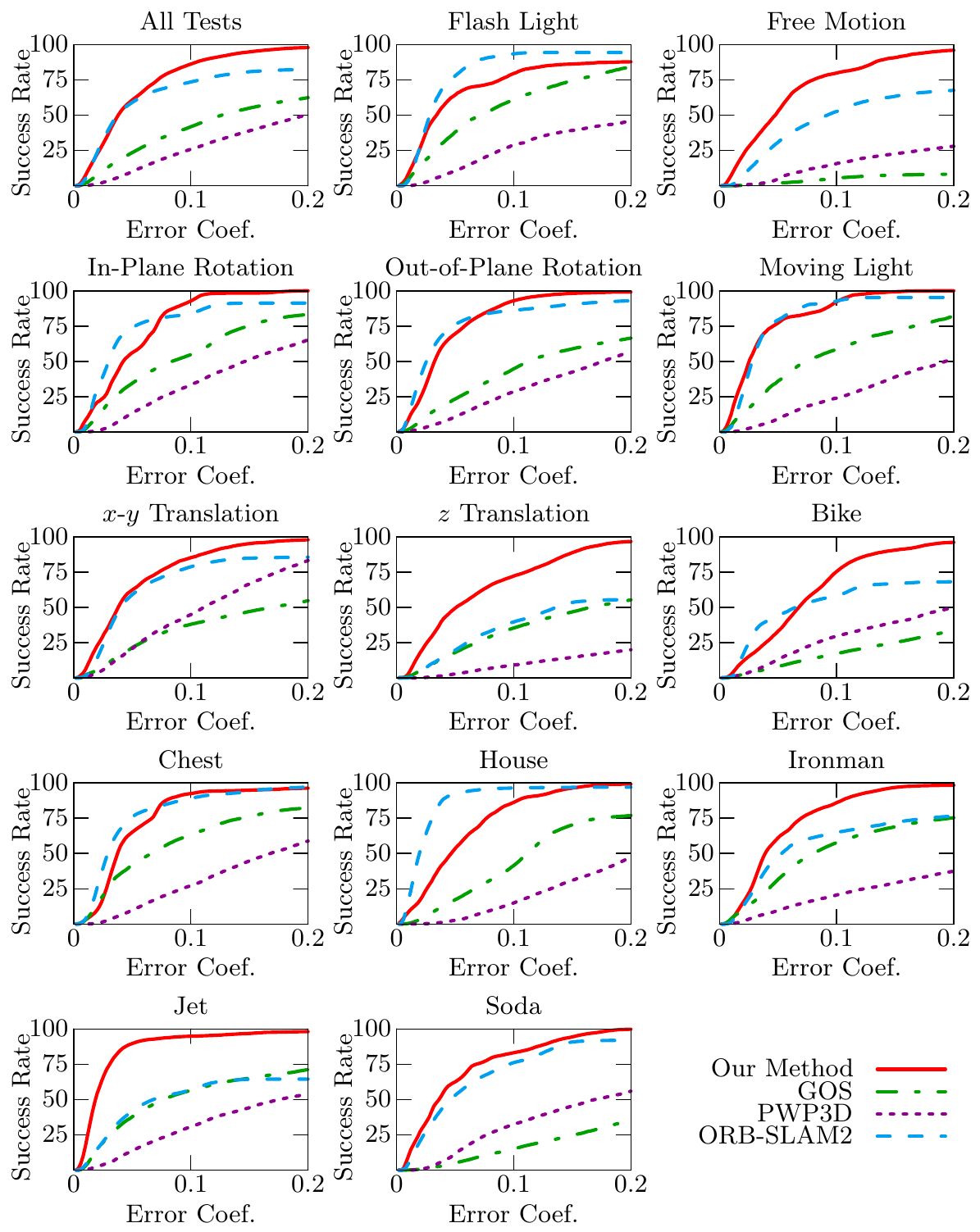}
\caption{
    Comparison of methods efficiency on various test groups
}
\label{fig:allGroupAndModelRes}
\end{figure}

Fig.~\ref{fig:allGroupAndModelRes} demonstrate the complete results of the
experiments on the described test groups.
Tables \ref{tab:results-by-models} and \ref{tab:results-by-groups} contain
corresponding AUC values.

\begin{table}
\begin{center}
    \caption{Comparison of AUC in tests with different objects}
    \label{tab:results-by-models}
    \begin{tabular}{p{0.2\textwidth} p{0.11\textwidth} p{0.1\textwidth}
    p{0.1\textwidth} p{0.1\textwidth} p{0.11\textwidth} p{0.1\textwidth}
    p{0.1\textwidth}}
    \hline
    Approach       & All tests   & Bike        & Chest       & House       & Ironman     & Jet         & Soda        \\
    \hline
    Our method     & {\bf 14.79} & {\bf 12.55} & 14.97       & 14.48       & {\bf 14.71} & {\bf 17.17} & {\bf 14.85} \\
    GOS            & 7.68        & 3.38        & 11.28       & 8.49        & 9.91        & 9.89        & 3.12        \\
    PWP3D          & 5.01        & 5.36        & 5.55        & 3.58        & 3.92        & 5.81        & 5.87        \\
    ORB-SLAM2      & 12.97       & 10.41       & {\bf 15.53} & {\bf 17.28} & 11.35       & 9.93        & 13.44       \\
    \hline
    \end{tabular}
\end{center}
\end{table}

\begin{table}
\begin{center}
    \caption{Comparison of AUC in tests with different tracking conditions}
    \label{tab:results-by-groups}
    \begin{tabular}{p{0.18\textwidth} p{0.09\textwidth}
    p{0.09\textwidth} p{0.12\textwidth} p{0.14\textwidth} p{0.11\textwidth}
    p{0.10\textwidth} p{0.1\textwidth}}
    \hline\noalign{\smallskip}
    Approach & Flash light        & Free motion         & In-plane rotation
             & Out-of-plane rot.  & Moving light        & $x$-$y$ transl.
             & {$z$ \newline transl.} \\
    \noalign{\smallskip}
    \hline
    \noalign{\smallskip}
    Our method     & 13.96       & {\bf 13.91} & {\bf 15.14} & {\bf 15.74} & {\bf 16.36} & {\bf 14.76} & {\bf 13.08} \\
    GOS            & 10.92       & 0.95        & 10.48       & 8.10        & 10.51       & 6.74        & 6.45        \\
    PWP3D          & 5.08        & 2.89        & 6.50        & 5.51        & 4.91        & 8.73        & 1.83        \\
    ORB-SLAM2      & {\bf 15.91} & 9.10        & 15.02       & 15.38       & 15.99       & 13.30       & 7.07        \\
    \hline
    \end{tabular}
\end{center}
\end{table}

\subsection{Dependency of Tracking Efficiency on Movement Speed}

\begin{figure}
\includegraphics{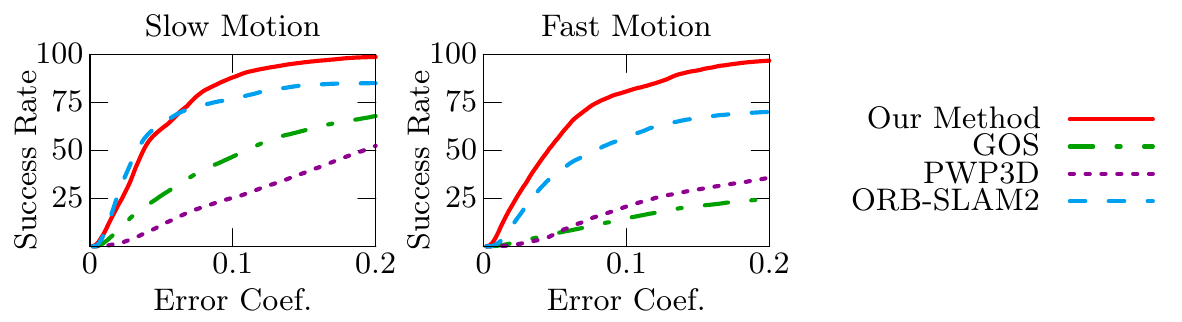}
\caption{
    Comparison of methods efficiency in tests grouped by speed
}
\label{fig:fastAndSlowRes}
\end{figure}

\begin{figure}
\includegraphics{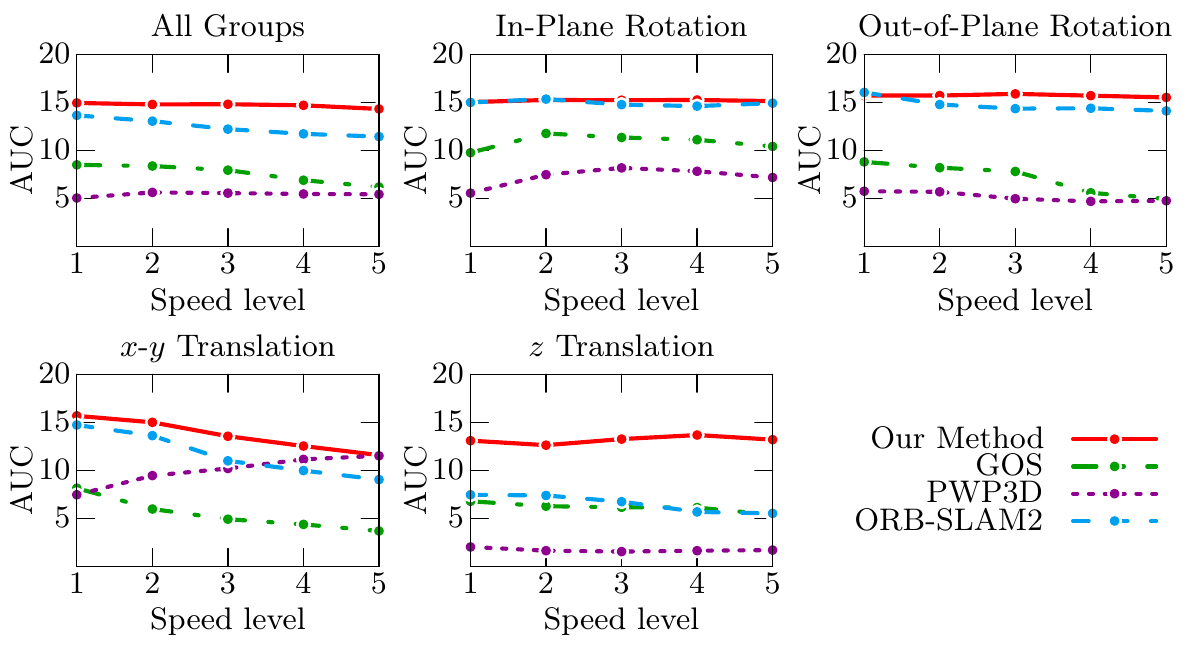}
\caption{
    Comparison of methods efficiency in tests containing different
    speed levels (level 5 stands for the fastest speed)
}
\label{fig:speedLevelsRes}
\end{figure}

To demostrate dependency of tracking efficiency on movement speed, we propose
two additional test groups:
\begin{itemize}
\item \term{Slow motion}. It contains translation and rotation tests with the
slowest speed level.
\item \term{Fast motion}. It contains translation and rotation tests with the
fastest speed level and free motion tests.
\end{itemize}
The results of these test groups are shown in Fig.~\ref{fig:fastAndSlowRes};
the corresponding AUC values are in
Table~\ref{tab:results-by-slowest-fastest}.
The dependency of tracking efficiency on all five movement speed levels
is shown in Fig.~\ref{fig:speedLevelsRes}.

\begin{table}
\begin{center}
   \caption{Comparison of AUC in the tests containing the slowest and the
   fastest motion}
   \label{tab:results-by-slowest-fastest}
   \begin{tabular}{p{0.2\textwidth} p{0.2\textwidth} p{0.2\textwidth}}
   \hline\noalign{\smallskip}
   Approach       & Slow motion & Fast motion \\
   \noalign{\smallskip}
   \hline
   \noalign{\smallskip}
   Our method     & {\bf 14.95} & {\bf 14.06} \\
   GOS            & 8.50        & 2.75        \\
   PWP3D          & 5.04        & 3.76        \\
   ORB-SLAM2      & 13.66       & 9.91        \\
   \hline
   \end{tabular}
\end{center}
\end{table}

\subsection{Run Time Performance}

Table~\ref{tab:perf-by-methods} demonstrates average frame processing times of
all tested methods.
Table~\ref{tab:sizes-of-models} contains sizes of tracked 3D~models.
Table~\ref{tab:perf-by-models} contains run time performance of our method and
GOS on different objects.

\begin{table}
\begin{center}
    \caption{Comparison of average frame processing time (ms)}
    \label{tab:perf-by-methods}
    \begin{tabular}{p{0.2\textwidth} p{0.15\textwidth} p{0.2\textwidth} p{0.2\textwidth}}
    \hline
    Our method  & GOS  & PWP3D  & ORB-SLAM2 \\
    \hline
    683         & 201  & 66     & 67 \\
    \hline
    \end{tabular}
\end{center}
\end{table}

\begin{table}
\begin{center}
    \caption{Sizes of tracked 3D~models}
    \label{tab:sizes-of-models}
    \begin{tabular}{p{0.15\textwidth} p{0.12\textwidth} p{0.12\textwidth} p{0.12\textwidth}
    p{0.12\textwidth} p{0.12\textwidth} p{0.12\textwidth}}
    \hline
    Value        & House & Soda & Bike   & Chest & Ironman & Jet    \\
    \hline
    Vertices     & 1448  & 3394 & 78562  & 14398 & 5763    & 88730  \\
    Faces        & 2594  & 6788 & 156950 & 28648 & 11496   & 176260 \\
    \hline
    \end{tabular}
\end{center}
\end{table}

\begin{table}
\begin{center}
    \caption{Our method and GOS run time performance on different objects}
    \label{tab:perf-by-models}
    \begin{tabular}{p{0.17\textwidth} p{0.11\textwidth} p{0.12\textwidth} p{0.12\textwidth}
    p{0.12\textwidth} p{0.11\textwidth} p{0.12\textwidth}}
    \hline
    Time (ms)  & Bike & Chest & House & Ironman & Jet  & Soda \\
    \hline
    Our method & 1097 & 358   & 550   & 427     & 1301 & 364  \\
    GOS        & 346  & 193   & 180   & 187     & 333  & 186  \\
    \hline
    \end{tabular}
\end{center}
\end{table}

\section{Model Contour Detection}\label{contour}

\newcommand{\Ico}{\mathcal{I}}
\newcommand{\IcoV}{V_\Ico}
\newcommand{\IcoF}{F_\Ico}

\newcommand{\MeshVxP}{\vect{p}}
\newcommand{\MeshVxQ}{\vect{q}}
\newcommand{\MeshVxA}{\vect{v}_1}
\newcommand{\MeshVxB}{\vect{v}_2}
\newcommand{\FaceA}{(\MeshVxA, \MeshVxQ, \MeshVxP)}
\newcommand{\FaceB}{(\MeshVxB, \MeshVxP, \MeshVxQ)}

\newcommand{\IcoVx}{\vect{u}}
\newcommand{\Invis}{S}

\newcommand{\InvisN}[1]{\Invis_{\IcoVx_{#1}}}
\newcommand{\IcoFace}{(\IcoVx_1, \IcoVx_2, \IcoVx_3)}

We have a 3D~model $\Mesh = (\MeshV, \MeshF)$, where $\MeshV \subset \Real^3$
is a finite set of model vertices, while $\MeshF$ is the set of triplets of
vertices defining model faces. The vertices are given in such a way that, when
one is looking at a front side of model face, they are arranged
counterclockwise. Current object pose is $\Pose$.

\paragraph{Face Visibility Pre-calculation}

Let us consider an icosphere $\Ico = \left( \IcoV, \IcoF \right)$, where $\IcoV
\subset \Real^3$ is a finite set of its vertices and  $\IcoF$ is a set of
vertex triplets that determine its faces. If the pivot points of the icosphere
$\Ico$ and the object model $\Mesh$ are aligned (let us assume that both are in
the origin of the current coordinate system), for each icosphere vertex $\IcoVx
\in \IcoV$, vector $-\IcoVx$ can be viewed as the model viewpoint direction.
Therefore, it is possible to obtain even orthographic projections of the object
model from all sides and, for each view direction $-\IcoVx$, to build the set
of invisible faces $\Invis_\IcoVx$.

\paragraph{Face Visibility in Current Pose}

Let us place the icosphere in pose $\Pose$ and determine which of its faces is
intersected by the ray connecting the center of the icosphere to the camera.
Let us label the vertices of that face $\IcoFace$. Let us label the set of the
model $\Mesh$ faces invisible from those vertices as $\Invis = \InvisN{1} \cap
\InvisN{2} \cap \InvisN{3}$. Most of the faces in $\Invis$ will be invisible in
the current pose. In most cases, the greater the $\abs{\IcoV}$, the fewer
visible faces in $\Invis$.

\paragraph{Contour and Sharp Edges Detection}

To detect contour and sharp edges of $\Mesh$, all model's edges formed by two
faces are reviewed. Let us consider edge $\MeshVxP \MeshVxQ$ and its adjacent
faces $\FaceA$ and $\FaceB$. The edge is considered sharp if both of its faces
are front and visible the angle between them is no greater then $\MinAngle$,
the edge is considered to be lying on the contour if one of its faces is front
and visible and the other is not (see Algorithm~\ref{algo:edge}).

\newcommand{\NormA}{\vect{n}_1}
\newcommand{\NormB}{\vect{n}_2}
\newcommand{\FrontA}{\vect{f}_1}
\newcommand{\FrontB}{\vect{f}_2}
\begin{algorithm}[H]
\caption{Edge type recognition}\label{algo:edge}
\begin{algorithmic}[1]
    \REQUIRE faces $\FaceA$ and $\FaceB$,
             set of invisible faces $\Invis$,
             angle $\MinAngle$
    \ENSURE type of the edge $\MeshVxP \MeshVxQ$
    \STATE $\NormA \leftarrow (\MeshVxQ - \MeshVxA) \times (\MeshVxP - \MeshVxA)$ \COMMENT{normal of the first edge}
    \STATE $\NormB \leftarrow (\MeshVxP - \MeshVxB) \times (\MeshVxQ - \MeshVxB)$ \COMMENT{normal of the second edge}
    \STATE $\FrontA \leftarrow \MeshVxA \cdot \NormA < 0$ \COMMENT{the first edge is front}
    \STATE $\FrontB \leftarrow \MeshVxB \cdot \NormB < 0$ \COMMENT{the second edge is front}
    \IF{$(\FrontA \land \neg \FrontB \land \FaceA \not\in \Invis)
            \lor (\FrontB \land \neg \FrontA \land \FaceB \not\in \Invis)$}
        \RETURN \Str{Contour}
    \ELSIF{$\FrontA \land \FrontB
            \land \FaceA \not\in \Invis \land \FaceB \not\in \Invis
            \land \pi - \arccos (\abs{\NormA \cdot \NormB}) > \MinAngle$}
        \RETURN \Str{Sharp}
    \ELSE
        \RETURN \Str{Other}
    \ENDIF
\end{algorithmic}
\end{algorithm}
\let\NormA\undefined
\let\NormB\undefined
\let\FrontA\undefined
\let\FrontB\undefined

\let\InvisN\undefined
\let\IcoFace\undefined

\let\IcoVx\undefined
\let\Invis\undefined

\let\MeshVxP\undefined
\let\MeshVxQ\undefined
\let\MeshVxA\undefined
\let\MeshVxB\undefined
\let\FaceA\undefined
\let\FaceB\undefined

\let\IcoV\undefined
\let\IcoF\undefined
\let\Ico\undefined